\documentclass{article}
\usepackage{spconf,amsmath,graphicx}
\usepackage{url}

\title{Transformer-based Models of Text Normalization for Speech Applications}
\name{Jae Hun Ro, Felix Stahlberg, Ke Wu, Shankar Kumar}
\address{Google Research\\
\texttt{\{jaero,fstahlberg,wuke,shankarkumar\}@google.com}}
\begin{document}
%
\maketitle
\begin{abstract}
Text normalization, or the process of transforming text into a consistent, canonical form, is crucial for speech applications such as text-to-speech synthesis (TTS). In TTS, the system must decide whether to verbalize "1995" as "nineteen ninety five" in "born in 1995" or as "one thousand nine hundred ninety five" in "page 1995". We present an experimental comparison of various Transformer-based sequence-to-sequence (seq2seq) models of text normalization for speech and evaluate them on a variety of datasets of written text aligned to its normalized spoken form. These models include variants of the 2-stage RNN-based tagging/seq2seq architecture introduced by \cite{zhang-etal-2019-neural}, where we replace the RNN with a Transformer in one or more stages, as well as vanilla Transformers that output string representations of edit sequences. Of our approaches, using Transformers for sentence context encoding within the 2-stage model proved most effective, with the fine-tuned BERT encoder yielding the best performance.

\end{abstract}


\begin{table*}
\centering
\small
\begin{tabular}{l@{\hspace{0.4em}}l}
Source &
\begin{tabular}{|@{\hspace{0.35em}}c@{\hspace{0.35em}}|@{\hspace{0.35em}}c@{\hspace{0.35em}}|@{\hspace{0.35em}}c@{\hspace{0.35em}}|@{\hspace{0.35em}}c@{\hspace{0.35em}}|@{\hspace{0.35em}}c@{\hspace{0.35em}}|@{\hspace{0.35em}}c@{\hspace{0.35em}}|@{\hspace{0.35em}}c@{\hspace{0.35em}}|@{\hspace{0.35em}}c@{\hspace{0.35em}}|@{\hspace{0.35em}}c@{\hspace{0.35em}}|@{\hspace{0.35em}}c@{\hspace{0.35em}}|@{\hspace{0.35em}}c@{\hspace{0.35em}}|@{\hspace{0.35em}}c@{\hspace{0.35em}}|@{\hspace{0.35em}}c@{\hspace{0.35em}}|@{\hspace{0.35em}}c@{\hspace{0.35em}}|@{\hspace{0.35em}}c@{\hspace{0.35em}}|@{\hspace{0.35em}}c@{\hspace{0.35em}}|@{\hspace{0.35em}}c@{\hspace{0.35em}}|@{\hspace{0.35em}}c@{\hspace{0.35em}}|@{\hspace{0.35em}}c@{\hspace{0.35em}}|@{\hspace{0.35em}}c@{\hspace{0.35em}}|@{\hspace{0.35em}}c@{\hspace{0.35em}}|@{\hspace{0.35em}}c@{\hspace{0.35em}}|}
\hline
\tiny{0} & \tiny{1} & \tiny{2} & \tiny{3} & \tiny{4} & \tiny{5} & \tiny{6} & \tiny{7} & \tiny{8} & \tiny{9} & \tiny{10} & \tiny{11} & \tiny{12} & \tiny{13} & \tiny{14} & \tiny{15} & \tiny{16} & \tiny{17} & \tiny{18} & \tiny{19} & \tiny{20} & \tiny{21}  \\
I & \  & l & i & v & e & \  & a & t & \ & 1 & 2 & 3 & \ & K & i & n & g & \ & A & v & e \\
\hline
\end{tabular} \\[1em]
Target (full) &
\begin{tabular}{|@{\hspace{0.35em}}c@{\hspace{0.35em}}|@{\hspace{0.35em}}c@{\hspace{0.35em}}|@{\hspace{0.35em}}c@{\hspace{0.35em}}|@{\hspace{0.35em}}c@{\hspace{0.35em}}|@{\hspace{0.35em}}c@{\hspace{0.35em}}|@{\hspace{0.35em}}c@{\hspace{0.35em}}|@{\hspace{0.35em}}c@{\hspace{0.35em}}|@{\hspace{0.35em}}c@{\hspace{0.35em}}|@{\hspace{0.35em}}c@{\hspace{0.35em}}|@{\hspace{0.35em}}c@{\hspace{0.35em}}|@{\hspace{0.35em}}c@{\hspace{0.35em}}|@{\hspace{0.35em}}c@{\hspace{0.35em}}|@{\hspace{0.35em}}c@{\hspace{0.35em}}|@{\hspace{0.35em}}c@{\hspace{0.35em}}|@{\hspace{0.35em}}c@{\hspace{0.35em}}|@{\hspace{0.35em}}c@{\hspace{0.35em}}|@{\hspace{0.35em}}c@{\hspace{0.35em}}|@{\hspace{0.35em}}c@{\hspace{0.35em}}|@{\hspace{0.35em}}c@{\hspace{0.35em}}|@{\hspace{0.35em}}c@{\hspace{0.35em}}|}
\hline
I & \  & l & i & v & e & \  & a & t & \  & one & twenty & three & \  & K & i & n & g & \  & Avenue \\
\hline
\end{tabular} \\[0.5em]
Target (edits) &
\begin{tabular}{|@{\hspace{0.35em}}c@{\hspace{0.35em}}|@{\hspace{0.35em}}c@{\hspace{0.35em}}|@{\hspace{0.35em}}c@{\hspace{0.35em}}|@{\hspace{0.35em}}c@{\hspace{0.35em}}|@{\hspace{0.35em}}c@{\hspace{0.35em}}|@{\hspace{0.35em}}c@{\hspace{0.35em}}|@{\hspace{0.35em}}c@{\hspace{0.35em}}|@{\hspace{0.35em}}c@{\hspace{0.35em}}|}
\hline
\texttt{pos10} & one & twenty & three & \texttt{pos13} & \texttt{pos19} & Avenue & \texttt{pos22} \\
\hline
\end{tabular} \\
\end{tabular}
\caption{Untokenized single-pass representations.}
\label{tab:single-pass-untokenized}
\end{table*}

\begin{table*}
\centering
\small
\begin{tabular}{ll}
Source &
\begin{tabular}{|@{\hspace{0.35em}}c@{\hspace{0.35em}}|@{\hspace{0.35em}}c@{\hspace{0.35em}}|@{\hspace{0.35em}}c@{\hspace{0.35em}}|@{\hspace{0.35em}}c@{\hspace{0.35em}}|@{\hspace{0.35em}}c@{\hspace{0.35em}}|@{\hspace{0.35em}}c@{\hspace{0.35em}}|@{\hspace{0.35em}}c@{\hspace{0.35em}}|@{\hspace{0.35em}}c@{\hspace{0.35em}}|@{\hspace{0.35em}}c@{\hspace{0.35em}}|@{\hspace{0.35em}}c@{\hspace{0.35em}}|@{\hspace{0.35em}}c@{\hspace{0.35em}}|@{\hspace{0.35em}}c@{\hspace{0.35em}}|@{\hspace{0.35em}}c@{\hspace{0.35em}}|@{\hspace{0.35em}}c@{\hspace{0.35em}}|@{\hspace{0.35em}}c@{\hspace{0.35em}}|@{\hspace{0.35em}}c@{\hspace{0.35em}}|@{\hspace{0.35em}}c@{\hspace{0.35em}}|@{\hspace{0.35em}}c@{\hspace{0.35em}}|@{\hspace{0.35em}}c@{\hspace{0.35em}}|@{\hspace{0.35em}}c@{\hspace{0.35em}}|@{\hspace{0.35em}}c@{\hspace{0.35em}}|@{\hspace{0.35em}}c@{\hspace{0.35em}}|}
\hline
I & \texttt{pos1} & l & i & v & e & \texttt{pos2} & a & t & \texttt{pos3} & 1 & 2 & 3 & \texttt{pos4} & K & i & n & g & \texttt{pos5} & A & v & e \\
\hline
\end{tabular} \\[0.5em]
Target (full) &
\begin{tabular}{|@{\hspace{0.35em}}c@{\hspace{0.35em}}|@{\hspace{0.35em}}c@{\hspace{0.35em}}|@{\hspace{0.35em}}c@{\hspace{0.35em}}|@{\hspace{0.35em}}c@{\hspace{0.35em}}|@{\hspace{0.35em}}c@{\hspace{0.35em}}|@{\hspace{0.35em}}c@{\hspace{0.35em}}|@{\hspace{0.35em}}c@{\hspace{0.35em}}|@{\hspace{0.35em}}c@{\hspace{0.35em}}|@{\hspace{0.35em}}c@{\hspace{0.35em}}|@{\hspace{0.35em}}c@{\hspace{0.35em}}|@{\hspace{0.35em}}c@{\hspace{0.35em}}|@{\hspace{0.35em}}c@{\hspace{0.35em}}|@{\hspace{0.35em}}c@{\hspace{0.35em}}|@{\hspace{0.35em}}c@{\hspace{0.35em}}|@{\hspace{0.35em}}c@{\hspace{0.35em}}|@{\hspace{0.35em}}c@{\hspace{0.35em}}|@{\hspace{0.35em}}c@{\hspace{0.35em}}|@{\hspace{0.35em}}c@{\hspace{0.35em}}|@{\hspace{0.35em}}c@{\hspace{0.35em}}|@{\hspace{0.35em}}c@{\hspace{0.35em}}|}
\hline
I & \texttt{pos1} & l & i & v & e & \texttt{pos2} & a & t & \texttt{pos3} & one & twenty & three & \texttt{pos4} & K & i & n & g & \texttt{pos5} & Avenue \\
\hline
\end{tabular} \\[0.5em]
Target (edits) &
\begin{tabular}{|@{\hspace{0.35em}}c@{\hspace{0.35em}}|@{\hspace{0.35em}}c@{\hspace{0.35em}}|@{\hspace{0.35em}}c@{\hspace{0.35em}}|@{\hspace{0.35em}}c@{\hspace{0.35em}}|@{\hspace{0.35em}}c@{\hspace{0.35em}}|@{\hspace{0.35em}}c@{\hspace{0.35em}}|}
\hline
\texttt{pos3} & one & twenty & three & \texttt{pos5} & Avenue \\
\hline
\end{tabular} \\
\end{tabular}
\caption{Tokenized single-pass representations.}
\label{tab:single-pass-tokenized}
\end{table*}

\begin{figure*}[t!]
\centering
\small
\includegraphics[width=1.0\textwidth]{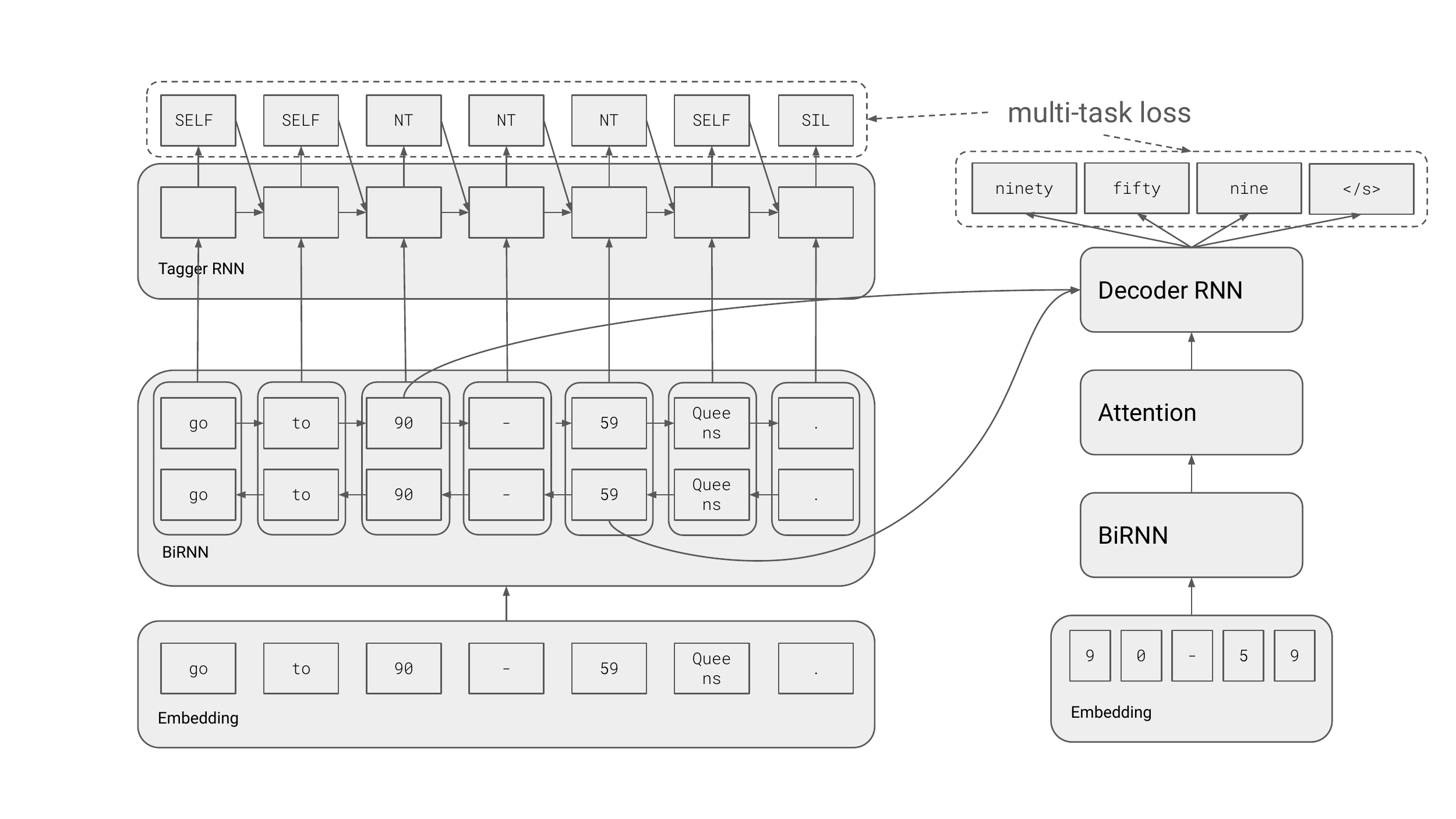}
\caption{Two stage model from \cite{zhang-etal-2019-neural}}
\label{fig:twostage}
\end{figure*}

\section{Introduction}
\label{sec:intro}
Text normalization is a key component for a wide range of speech and language processing applications. However, depending on the particular application, the requirements for text normalization systems can vary greatly. We examine text normalization in the context of text-to-speech (TTS) synthesis, where written text is read aloud. In this scenario, numbers (e.g. \texttt{123}), which might typically be mapped to some unified symbol for all numbers (e.g. \texttt{N}) during pre-processing, must instead be translated into their spoken form (e.g. \texttt{one hundred twenty three}) as part of the front-end of the TTS system. Numbers are just one instance of \emph{non-standard words} \cite{SPROAT2001287} that have different pronunciations depending on the context. We can further categorize these non-standard words into \emph{semiotic classes} \cite{taylor_2009}, such as measures, dates, and URLs.

Early works on text normalization in speech were largely based on hard-coded rules \cite{allen1987text,sproat_1996,sproat1997multilingual} and later, a combination of hand-written grammars \cite{ebden_sproat_2015} and machine learning for specific semiotic classes: \cite{Sproat2014ApplicationsOM} for letter sequences, \cite{roark-sproat-2014-hippocratic-short} for abbreviations, \cite{gorman-sproat-2016-minimally} for cardinals, and \cite{zhang-etal-2020-semi-short} for URLs.
More recently, there has been increased interest in neural models due to their promise of better accuracy for less maintenance compared to grammars \cite{sproat2016rnn,Sproat2017AnRM,yolchuyeva2018,zhang-etal-2019-neural,mansfield2019,finch-etal-2021-apple,tyagi-etal-2021-proteno}.

\cite{sproat2016rnn} first presented a variety of RNN-based architectures for text normalization along with an open-sourced corpus of written text paired with its spoken form. A subsequent study \cite{Sproat2017AnRM} concentrated on the attention-based sequence-to-sequence (seq2seq) model based on \cite{chan2016las} that outperformed the other models along with a finite-state transducer (FST)-based filter to mitigate the unrecoverable errors produced by the RNN alone (e.g. \texttt{50 ft} $\rightarrow$ \texttt{fifty inches}). \cite{zhang-etal-2019-neural} further improved upon this work by presenting neural architectures with increased accuracy and efficiency and covering grammars largely learned from data.

Originally presented in the context of machine translation (MT), the Transformer \cite{vaswani2017} relies on attention mechanisms to eliminate recurrence entirely.
The architecture consists of an encoder with self-attention and an auto-regressive decoder with self-attention that also attends to the encoder.
While \cite{zhang-etal-2019-neural} reported worse sentence error rates with the full Transformer relative to their model, they sought to showcase the issues of treating text normalization as a pure MT task rather than evaluate the effectiveness of the architecture.
Transformers have since been extended far beyond MT tasks, with both the encoder and decoder proving useful independently \cite{devlin-etal-2019-bert-short,Radford2018ImprovingLU,Brown2020GPT3}.
One especially salient example is BERT \cite{devlin-etal-2019-bert-short} which makes use of bidirectional pre-training for language representations using a masked language modeling objective function. These pre-trained BERT models have often been shown to improve numerous downstream tasks, including, but not limited to, question answering, text classification, and summarization.

We present a survey of several Transformer-based models of text normalization for speech and examine two separate datasets derived from English Wikipedia: \emph{Standard} \cite{sproat2016rnn} data mined in 2016, run through the Google TTS Kestrel text normalization system \cite{ebden_sproat_2015}, and released on GitHub \footnote{\url{https://github.com/rwsproat/text-normalization-data}}, and \emph{Manual} annotated data mined from Wikipedia that was sent to human raters for manual annotation.
We describe our models in Section~\ref{sec:models} and present training and evaluation results in Section~\ref{sec:results}.

\begin{table*}[t]
\centering
\begin{tabular}{lcc}
\hline
& Test SER \% & \\\cline{2-3}
System & Standard & Manual \\
\hline
\cite{zhang-etal-2019-neural} & $2.25$ & $-$ \\
\cite{zhang-etal-2019-neural}* & $1.80$ & $-$ \\
\cite{stahlberg-kumar-2020-seq2edits-short} & 1.36 & $-$ \\
\hline
Untokenized single-pass (full) & $2.32$ & $7.55$ \\
Untokenized single-pass (edits) & $2.12$ & $8.00$ \\
Tokenized single-pass (full)* & $2.00$ & $6.16$ \\
Tokenized single-pass (edits)* & $2.04$ & $5.65$ \\
\hline
RNN (base) & $1.99$ & $7.18$ \\
RNN (large) & $2.15$ & $8.09$ \\
Transformer encoder & $1.73$ & $6.49$ \\
Transformer encoder (seq2seq) & $2.15$ & $7.40$ \\
BERT (fine-tune) & $1.42$ & $5.59$ \\
BERT (freeze) & $2.68$ & $7.96$ \\
\hline
\end{tabular}
\caption{Test sentence error rates for each dataset. * requires golden tokenized input.
}
\label{tab:tft-ser}
\end{table*}

\section{Models}
\label{sec:models}
\subsection{Single-pass models}

In this section, we explore vanilla seq2seq Transformers as proposed by \cite{vaswani2017} that can be trained with normal cross-entropy loss and used with standard single-pass beam search.
Our first baseline maps the character sequence of the full input sentence to its verbalization (a mix of characters and full words for normalized input), i.e.\ {\em Source} to {\em Target (full)} in Table \ref{tab:single-pass-untokenized}. To avoid the need for copying unchanged characters, we also experiment with an edit-based representation that uses special \texttt{pos*} tokens to identify spans in the source sequence, shown as {\em Target (edits)} in Table \ref{tab:single-pass-untokenized}. In this example, the tokens \texttt{pos10} and \texttt{pos13} denote the replacement from the 10th-13th character in the input with {\em one twenty three}. This pattern can be repeated in the output sequence to represent non-contiguous replacements. If we assume that the tokenization is known (Table \ref{tab:single-pass-tokenized}), we can use the \texttt{pos*} position tokens in the input to denote token boundaries in the source sequence. The seq2seq model that also outputs \texttt{pos*} markers at token boundaries ({\em Target (full)} in Table \ref{tab:single-pass-tokenized}) can use this information to keep track of the alignment between source and target. Edit-based output representations for tokenized inputs are simpler because the character span to replace can be identified by a single token index (e.g. \texttt{pos3} in the {\em Target (edits)} example in Table \ref{tab:single-pass-tokenized}) rather than a pair of start/end character positions (e.g. \texttt{pos10} and \texttt{pos13} in the {\em Target (edits)} example in Table \ref{tab:single-pass-untokenized}).

\subsection{Stacked tagging and contextual models}

The best RNN-based models reported by \cite{zhang-etal-2019-neural} use a 2-stage neural architecture (Figure~\ref{fig:twostage}) specialized for text normalization. The model is trained with a multi-task objective and tokenization levels differ across various stages.
In this section, we present a series of modifications to the 2-stage architecture.
The first stage, \emph{semiotic class tokenization}, jointly segments and coarsely classifies tokens as trivial (\texttt{<self>}), silent (\texttt{sil}), or non-trivial (NT). In the second stage, a seq2seq model predicts the normalization for non-trivial tokens.
Below, we focus on the sentence context encoder because both stages share its hidden states.

\emph{RNN (base)} 2-stage multi-task model from \cite{zhang-etal-2019-neural}.

\emph{RNN (large)} increases the size of \emph{RNN} for a more fair comparison against Transformers by adding layers to the sentence context encoder.

\emph{Transformer encoder} replaces the sentence context encoder (block labeled "BiRNN" for bidirectional RNN in Figure~\ref{fig:twostage}) with the base Transformer encoder. This can potentially enhance the semiotic class tokenization by using long-term context which is not handled as well by recurrent models.

\emph{Transformer encoder (seq2seq)} replaces the seq2seq encoder for token verbalization with the base Transformer encoder as a comparison point.

\emph{BERT (fine-tune)} replaces the sentence context encoder with a pre-trained BERT word-piece \cite{sennrich-etal-2016-neural-short,45610} model. Given the effectiveness of fine-tuning BERT for down-stream tasks \cite{devlin-etal-2019-bert-short}, it could also improve text normalization.

\emph{BERT (freeze)} matches the previous but freezes the pre-trained model as freezing has been shown to accelerate training and potentially be more stable than fine-tuning \cite{mosbach2021on,lee2019elsa}.

\begin{table}
\centering
\begin{tabular}{lcc}
\hline
& Test SER \% & \\\cline{2-3}
System & Standard & Manual \\
\hline
RNN (base) & & \\
\hspace{5mm}base & $1.99$ & $7.18$ \\
\hspace{5mm}+golden & $0.77$  & $2.74$ \\
BERT (fine-tune) & & \\
\hspace{5mm}base & $1.42$ & $5.59$ \\
\hspace{5mm}+golden & $0.69$ & $2.59$ \\
\hline
\end{tabular}
\caption{Test sentence error rates for the 2-stage models with ("+golden") and without ("base") the golden semiotic class tokenization.}
\label{tab:tft-rnn-golden-ser}
\end{table}

\section{Results}
\label{sec:results}

\begin{table*}[t]
\centering
\begin{tabular}{lllll}
Input & Reference & RNN (base) & BERT (fine-tune) \\
\hline
\texttt{[AAUS]} & \texttt{a a u s} & \texttt{aaus} & \texttt{a a u s} \\
\texttt{July [93]} & \texttt{ninety three} & \texttt{ninety third} & \texttt{ninety three} \\
\texttt{Final Fantasy [X]} & \texttt{ten} & \texttt{x} & \texttt{ten} \\
\texttt{[CHARLES]} & \texttt{charles} & \texttt{c h a r l e s} & \texttt{charles} \\
\hline
\texttt{chief [ideologue]} & \texttt{ideologue} & \texttt{homolog} & \texttt{homolog} \\
$[\frac{7}{8}]$ \texttt{inch} & \texttt{seven eighths} & \texttt{one dollar} & \texttt{five eighth} \\
\hline
\texttt{[TIME] Magazine} & \texttt{t i m e} & \texttt{time} & \texttt{time} \\
\texttt{Cha Seung [pyo's]} & \texttt{p y o's} & \texttt{pyo's} & \texttt{pyo's} \\
\end{tabular}
\caption{Standard dataset side-by-side of RNN (base) and BERT (fine-tune). \texttt{[]} marks the target token in context. The top section lists examples corrected by BERT (fine-tune), the middle, uncorrected errors, and the bottom, mistakes in the reference.}
\label{tab:tft-sxs}
\end{table*}

Table~\ref{tab:tft-ser} reports sentence error rates (SER) for each model and dataset.
Comparing our models and the original 2-stage model from \cite{zhang-etal-2019-neural}, we find that using Transformers in sentence context encoding is most effective, with the BERT (fine-tune) yielding the best performance.
This could be because the model is able to leverage the underlying language representations from BERT to focus more on the interesting, non-trivial cases compared to training from scratch, especially when the training data size is as small as it is here.
The single-pass models are less robust to the small training data size because not all position tokens \texttt{pos*} are seen often enough in training data to learn reliably.
While our models fall slightly short of the performance obtained by the sequence editing approach from \cite{stahlberg-kumar-2020-seq2edits-short}, our focus is primarily on comparing various Transformer models for text normalization and less on surpassing the state-of-the-art.

Table~\ref{tab:tft-rnn-golden-ser} provides a more detailed breakdown of the improvements obtained using BERT. We find that mistakes in the first stage semiotic class tokenization (i.e. classifying trivial as non-trivial and vice-versa) account for a majority of the errors. Given the golden semiotic class tokenization, the error rate is reduced by $61\%$ for RNN (base). After applying BERT, we see reductions in both the error rates and the gap between them. This observation is further supported when we examine the actual improvements and errors.

Table~\ref{tab:tft-sxs} provides a side-by-side comparing the 2-stage models. 
Given the input \texttt{AAUS}, RNN (base) incorrectly predicts \textit{trivial}, resulting in \texttt{aaus}, whereas BERT (fine-tune) correctly predicts \textit{non-trivial}, resulting in a letter sequence \texttt{a a u s}. On the other hand, both models misclassify \texttt{ideologue} as \textit{non-trivial} possibly due to normalized British English training examples, such as \texttt{analogue} $\rightarrow$ \texttt{analog}, resulting in the unrecoverable \texttt{homolog}.\footnote{In theory, the seq2seq model could correctly predict the identity mapping but this hardly occurs in the training data and there is no guarantee that the output vocabulary covers it.} While this unrecoverable error is a result of \emph{trivial} / \emph{non-trivial} misclassification, it is worth mentioning that these mistakes are possible even with the correct classification: e.g. \texttt{$\frac{7}{8}$ inch} $\rightarrow$ \texttt{five eighth inch}.

While the above examples are genuine errors, the Standard dataset is also known to contain reference errors \cite{sproat2016rnn} because it was automatically generated with Google's Kestrel text normalization system \cite{ebden_sproat_2015}: \texttt{TIME magazine} $\rightarrow$ \texttt{t i m e magazine}.
It may be tempting to view manual annotation as the solution, but even with human raters, mistakes still occur and can be harder to track down due to rater inconsistency: given \texttt{UEFA Champions League}, some raters may provide \texttt{uefa champions league}, while others provide \texttt{u e f a champions league}.

\section{Conclusion}
We presented a survey of Transformer-based models for text normalization for speech and evaluated their performance on different datasets, with our BERT fine-tuning approach yielding the most improvement. While we did achieve good overall accuracy, we also showed that our approaches are still vulnerable to unrecoverable errors.
We hope that our work inspires more investigation into text normalization for speech and provides additional evidence that simply switching architectures, using pre-training recipes, or adding labeled data will not completely solve the text normalization problem.


\bibliographystyle{IEEEbib}
\bibliography{anthology,custom}

\end{document}